\documentclass{article}
\usepackage{arxiv}
\usepackage{booktabs,makecell, multirow, tabularx}
\usepackage[utf8]{inputenc} 
\usepackage[T1]{fontenc}    
\usepackage{hyperref}       
\usepackage{url}            
\usepackage{booktabs}       
\usepackage{amsfonts}       
\usepackage{nicefrac}       
\usepackage{microtype}      
\usepackage{lipsum}
\usepackage{graphicx}
\usepackage{threeparttable}
\usepackage{amsmath,amsfonts}
\usepackage[caption=false,font=normalsize,labelfont=sf,textfont=sf]{subfig}
\graphicspath{ {./images/} }

\title{FMOcc: TPV-Driven Flow Matching for 3D Occupancy Prediction with Selective State Space Model}

\author{
 Jiangxia Chen \\
  School of Artificial Intelligence\\
  Chongqing University of Technology\\
  Chongqing, China 401135\\
  \texttt{jiangxiachen0@gmail.com} \\
   \And
 Tongyuan Huang  \\
  School of Artificial Intelligence\\
  Chongqing University of Technology\\
  Chongqing, China 401135\\
  \texttt{tyroneh@cqut.edu.cn} \\
  \And
 Ke Song \\
  School of Intelligent Systems Engineering \\
  Sun Yat-sen University\\
  Shenzhen, China  528406\\
  \texttt{songk9@mail2.sysu.edu.cn} \\
}

\begin{document}
\maketitle
\begin{abstract}
3D semantic occupancy prediction plays a pivotal role in autonomous driving. However, inherent limitations of few-frame images and redundancy in 3D space compromise prediction accuracy for occluded and distant scenes. Existing methods enhance performance by fusing historical frame data, which need additional data and significant computational resources.  To address these issues, this paper propose FMOcc, a Tri-perspective View (TPV) refinement occupancy network with flow matching selective state space model for few-frame 3D occupancy prediction. Firstly, to generate missing features, we designed a feature refinement module based on a flow matching model, which is called Flow Matching SSM module (FMSSM). Furthermore, by designing the TPV SSM layer and Plane Selective SSM (P$\text{S}^{3}$M), we selectively filter TPV features to reduce the impact of air voxels on non-air voxels, thereby enhancing the overall efficiency of the model and prediction capability for distant scenes. Finally, we design the Mask Training (MT) method to enhance the robustness of FMOcc and address the issue of sensor data loss. Experimental results on the Occ3D-nuScenes and OpenOcc datasets show that our FMOcc outperforms existing state-of-the-art methods. Our FMOcc with two frame input achieves notable scores of 43.1\% RayIoU and 39.8\% mIoU on Occ3D-nuScenes validation, 42.6\% RayIoU on OpenOcc with 5.4 G inference momory and  330ms inference time.
\end{abstract}


\section{Introduction}
In recent years, there has been a notable transition in the field of autonomous driving technology, with a shift from LiDAR-based multi-modal fusion to image-based vision-centric perception. This transition has been driven primarily by the potential to reduce reliance on costly LiDAR sensors \cite{OS, xiong2022road}. The occupancy network has emerged as a cornerstone in vision-centric methodologies to capture the dense 3D structure of the real world \cite{liang2024suprnet, HighPosition,wang2024opus,liao2025stcocc,yang2024adaptiveocc,ouyang2024linkocc}. This developing field of perception technology aims to infer the occupied state of each voxel in a voxelized world instead of predicting 3D box \cite{zhu2024drop}. It demonstrates a robust capacity for generalization, enabling its application to open-set objects, irregularly shaped vehicles and specialized road structures \cite{wang2024panoocc, CommunEng}. 

Despite there is more and more research being done on the design of 3D occupancy networks, a prevalent challenge in this field is training a robust network that can accommodate occlusion cases and maintain satisfactory predictive performance, even in the event of feature loss. Co-Occ \cite{pan2024co} combines explicit 3D feature representation with implicit volume-based regularization to enhance inter-modal interaction and improve the fused volumetric representation. Although Co-Occ enhances the robustness by fusing the LiDAR features and camera features, the performance of this method still decreases when the features loss. Furthermore, the occlusion cases are still difficult.

The diffusion model \cite{song2020denoising, ho2020denoising} demonstrates its powerful generative capabilities. Several recent studies have applied diffusion models to many perceptual tasks, such as object detection \cite{chen2023diffusiondet}, semantic segmentation \cite{wolleb2022diffusion} and depth estimation \cite{saxena2023monocular}. Furthermore, diffusion models are increasingly used in 3D perception tasks, such as 3d occupancy prediction \cite{wang2024occgen}. Although OccGen \cite{wang2024occgen} employs a "noise-to-occupancy" generative paradigm, whereby noise originating from a random 3D Gaussian distribution is progressively inferred and eliminated. OccGen does not improve for feature loss cases or occlusion cases. Moreover, diffusion models often suffer from mode collapse issues, where the generated outputs tend to converge to a limited set of patterns rather than fully exploring the diversity of the data distribution. This limitation becomes particularly evident when dealing with high-dimensional data spaces, such as in 3D point clouds or volumetric representations.

The objective of this paper is to enhance the robustness and generative power of generative-based occupancy prediction while simultaneously reducing both computational efficiency and computational costs. The use of 3D voxel as input to the generative model and the use of Transformer as model to the diffusion model \cite{wang2024occgen} take up a large amount of computational resources. The recent emergence of state space models (SSMs), such as Mamba \cite{gu2023mamba, zhu2024vision}, has led to the development of more efficient solutions for long-range modeling. In view of this, we are exploring the potential of state space models for improving occupancy prediction tasks.

In this work, we present 3D semantic occupancy prediction with the proposed Flow Matching SSM (FMSSM), which we refer to as FMOcc. The generative power of the flow matching model enables our FMOcc to effectively address 3D semantic occupancy prediction in the cases of feature loss and occlusion.  This approach not only significantly reduces the computational complexity by eliminating the need for numerous iterative steps but also enhances the model's ability to capture complex dependencies within the data compared to diffusion model. To enhance the efficacy of the flow matching model in 3D semantic occupancy prediction network, our proposed FMOcc employs tri-perspective view (TPV) \cite{huang2023tri} features in instead of 3D voxel features as the input for the flow matching model. Following the conversion of the TPV features to 1D features, our SSMs-based flow matching model is capable of effectively handling long-range features, thereby benefiting global modeling and the linear computation complexity of SSMs. By integrating this generative approach with the global modeling proficiencies of SSMs, we enhance capabilities of FMOcc to encompass occlusion reasoning and comprehension of distant scenes, especially in complex driving scenes. Comprehensive experiments are conducted using the challenging Occ3D-nuScenes and OpenOcc dataset to validate our methodology. The mIoU, RayIoU and IoU are used as the metrics to evaluate the performance of our FMOcc. Our FMOcc consistently outperforms state-of-the-art algorithms in the Occ3D-nuScenes and OpenOcc benchmarks.  

The contributions of our paper are summarized as follows:
\begin{itemize}
	\item[$\bullet$] We propose FMOcc for 3D semantic occupancy prediction which is based on flow matching models and selective SSMs.
	\item[$\bullet$] The proposed FMOcc significantly outperforms previous methods in term of 3D semantic occupancy prediction on the Occ3D-nuScenes and OpenOcc datasets. 
	\item[$\bullet$] FMOcc is not limited to predict the 3D occupancy in the common cases and can be effective used in the cases of low-light scenes, distant objects and occlusion situations.
	\item[$\bullet$] Ablation experiments demonstrate that FMOcc is a consequence of framework innovation, and its core components exert a beneficial influence on the 3D occupancy prediction framework.
\end{itemize}

\section{Related Work}
\subsection{3D Semantic Occupancy Prediction}
3D semantic occupancy prediction aims to estimate the surrounding environment by combining geometric and semantic information. MonoScene \cite{cao2022monoscene} proposes the first 3D semantic scene reconstruction (SSC) framework that infers dense geometry and semantic information of the scene from a single RGB image. VoxFromer \cite{li2023voxformer}, a transformer-based two-stage semantic scene completion framework, is capable of predicting voxel occupancy and semantic information within a spatial context solely from 2D images. OccFormer \cite{zhang2023occformer}, a dual-path transformer network, is designed to efficiently handle semantic occupancy prediction of 3D volumes. By decomposing the intensive 3D processing into local and global transformation paths along the horizontal plane, long-range, dynamic, and efficient encoding of camera-generated 3D voxel features is conducted. COTR \cite{ma2024cotr} introduces a compact occupancy transformer which includes a geometry-aware occupancy encoder and a semantic-aware group decoder. GEOcc \cite{tan2024geocc} geometrically enhances the view transformation process by integrating implicit and explicit depth modeling. However, previous camera-only occupancy prediction methods are based on discriminative modeling which is lack of the robustness and generalizability. 

\subsection{Generative Models for 3D Perception}
Diffusion models have been recently adopted in computer vision tasks, such as text-to-3D \cite{lin2023magic3d,poole2022dreamfusion} and image-to-3D \cite{liu2023zero}. Furthermore, diffusion models have also widely used in perception tasks such as 3D object detection \cite{le2024diffuser}, 3D semantic segmentation \cite{zou2024diffbev}, depth estimation \cite{ji2023ddp} and 3D occupancy prediction \cite{wang2024occgen}. For 3D occupancy prediction task, OccGen \cite{wang2024occgen} uses diffusion model to refine the multi-modal occupancy in a coarse-to-fine manner. OccGen employs the voxel as an input to the transformer in the diffusion model. However, the use of voxel as input to the diffusion model has the effect of reducing the computational efficiency of the model, and OccGen just use the diffusion model to refine the occupancy. In contrast, Flow Matching offers several distinct advantages over traditional diffusion models. By formulating the generation process as an optimization problem that matches the flow of data distributions, it enables more direct and efficient mapping between latent spaces and target data. This approach not only significantly reduces the computational complexity by eliminating the need for numerous iterative steps but also enhances the model's ability to capture complex dependencies within the data.
\begin{figure*}[h]
	\centering
	\includegraphics[width=1.0\textwidth]{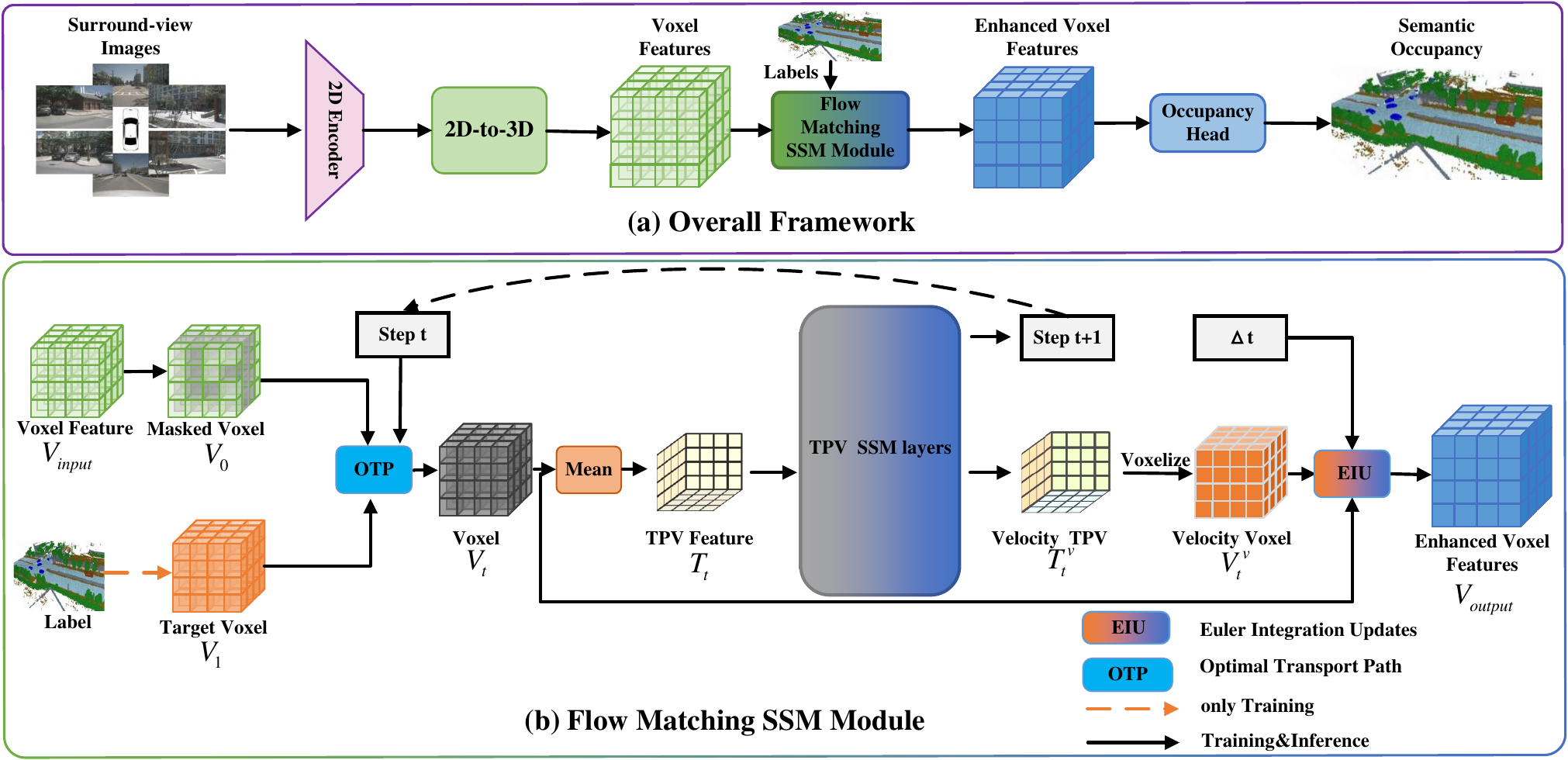}
	\caption{Schematic overview of our FMOcc. Given surround-view images, we first use 2D encoder to extract the camera features. LSS \cite{huang2021bevdet} is utilized to transform the 2D features to 3D voxel features. Then, 3D voxel features are defined as the input of Flow Matching SSM module to produce the enhanced 3D voxel features. Finally, the enhanced 3D voxel features are fed to the occupancy head, producing semantic occupancy predictions. (a) Demonstrates the overall framework of FMOcc. (b) Shows the structure of Flow Matching SSM module.}
	\label{fig_overall_framework}
\end{figure*}

\begin{figure*}[h]
	\centering 
	\includegraphics[width=\textwidth]{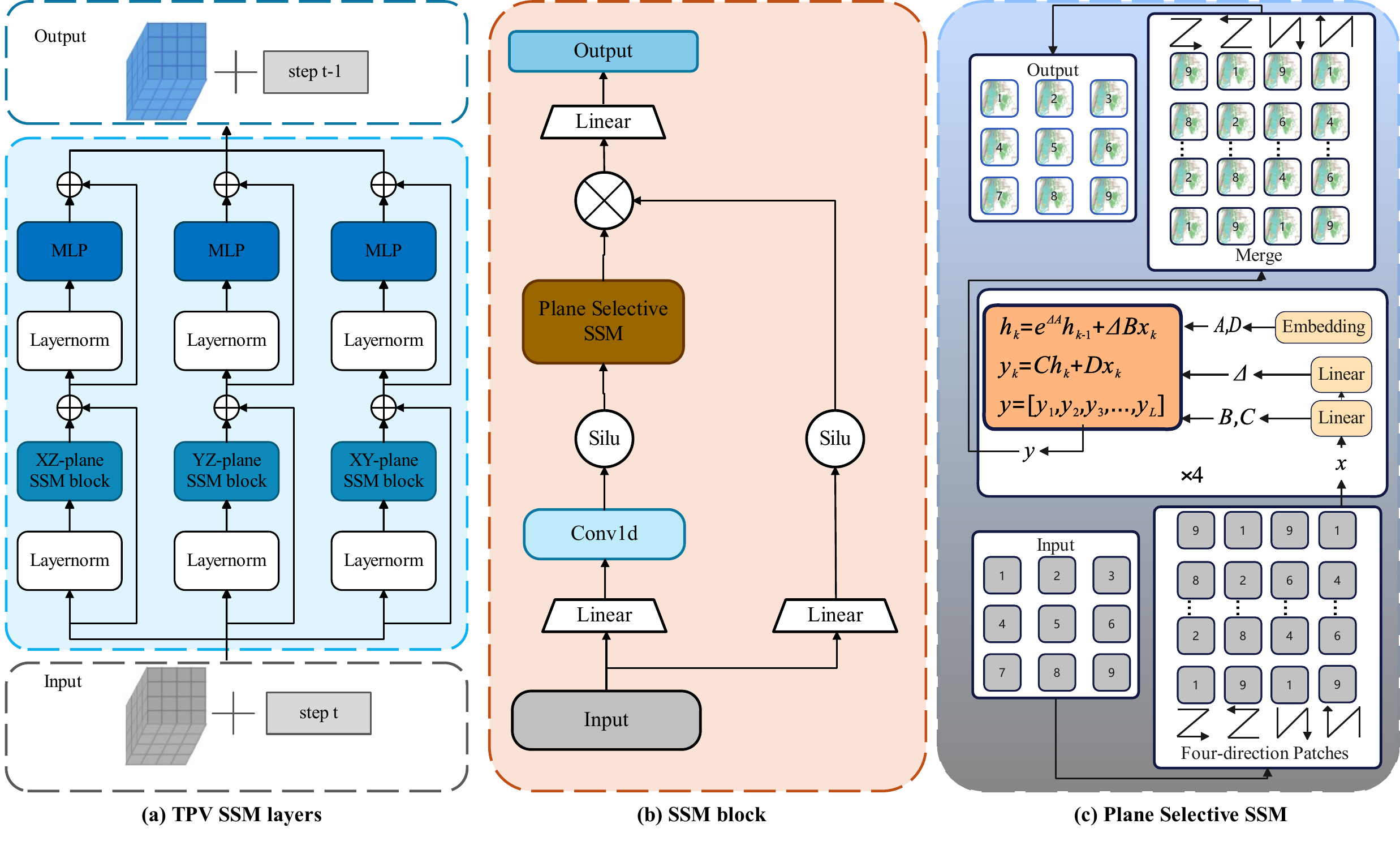}
	\caption{Schematic overview of our TPV SSM layers. (a) The inputs of TPV features $T_{t}$ and step t are processed by three parallel layers to obtain the output of $T_{t-1}$ and step t-1. (b) Demonstrates framework pf the SSM blocks of three planes. (c) Shows structure of the proposed Plane Selective SSM ($PS^{3}M$).}
	\label{tpv_ssm_layers}
\end{figure*}
Our method aims to leverage the efficiency of tri-plane features and the generative property of flow matching model to mitigate the lack of features which are extracted from images. Furthermore, our method can improves the performance of 3D occupancy prediction for occluded objects.

\subsection{State Space Models for 3D Perception}
The transformer \cite{vaswani2017attention} has changed the computer vision industry, but its quadratic computation complexity presents a challenge. The state space model series (SSMs), including Mamba \cite{gu2023mamba}, are becoming increasingly prominent. Mamba distinguishes itself through the incorporation of selective mechanisms that efficiently capture long-range dependencies and handle large-scale data within linear time complexity. Recently, some works \cite{zhu2024vision, huang2024localmamba} have introduced Mamba to computer vision. In the field of 3D perception, Voxel-Mamba \cite{zhang2024voxel} use Mamba instead of transformer to process voxel for 3D object detection. OccMamba \cite{li2024occmamba} proposes hierarchical Mamba encoder and decoder to improve the performance of 3D occupancy rediction. Due to the efficiency of SSM, we design SSM-based flow matching model for 3D occupancy prediction compared with OccGen. 

\section{Proposed Method}
In this section, we outline how the proposed FMOcc is utilized to enhance the generative power and efficiency of model facing feature loss and occlusion for 3D semantic occupancy prediction task. In the following sections, we first provide a brief review of flow matching models and SSMs in Section \ref{Preliminary}. Then, we present the framework overview of FMOcc in Section \ref{Framework Overview}. Thereafter, we provide a detailed explanation of the Flow Matching SSM Module (FMSSM) which is the core component of FMOcc in Section \ref{Flow Matching SSM Module}. Additionally, the Tri-perspective views SSM (TPV-SSM) layers are demonstrated in details in Section \ref{TPV-SSM}. Finally, we introduce a novel training process which is defined as Mask Training (MT) in Section \ref{Mask Training}.

\subsection{Preliminary}\label{Preliminary}
Our proposed FMOcc is based on flow matching models and SSMs. Thus, this section provides an overview of the flow matching models and SSMs.

\subsubsection{Flow Matching Model} 
In contrast to diffusion models, which center on learning to reverse the progressive introduction of noise over time for data recovery, flow matching\cite{lipman2022flow} emphasizes learning invertible transformations that establish mappings between data distributions. Let $\pi_0$ represent a simple distribution, commonly the standard normal distribution $p(x) = \mathcal{N}(x|0, I)$, and let $\pi_1$ stand for the target distribution. Within this framework, rectified flow\cite{liu2022flow} leverages a straightforward yet effective approach to construct paths via optimal transport\cite{mccann1997convexity} displacement, and we adopt this as our Flow Matching method.

Given $x_0$ sampled from $\pi_0$, $x_1$ sampled from $\pi_1$, and $t \in [0, 1]$, the path from $x_0$ to $x_1$ is defined as a straight line. This implies that the intermediate state $x_t$ is given by $(1 - t)x_0 + tx_1$, with the direction of the intermediate state consistently following $x_1 - x_0$. By building a neural network $v_\theta$ to predict the direction $x_1 - x_0$ based on the current state $x_t$ and the time step $t$, we can derive a path from the initial distribution $\pi_0$ to the target distribution $\pi_1$ by optimizing the loss between $v_\theta(x_t, t)$ and $x_1 - x_0$. This process can be formalized as follows:

\begin{equation}
v_\theta(x_t, t) \approx \mathbb{E}_{x_0 \sim \pi_0, x_1 \sim \pi_1} [v_t | x_t]
\end{equation}

\begin{equation}
\mathcal{L}(\theta) = \mathbb{E}_{x_0 \sim \pi_0, x_1 \sim \pi_1} \left[ \left\| v_\theta(x_t, t) - (x_1 - x_0) \right\|_2 \right]
\end{equation}

\subsubsection{State Space Models}
The state space modeks (SSMs) are inspired by the control theory. It can handling sequences features effectively. For discrete inputs, the mathematical formulation of SSMs is presented as follows.:
\begin{equation}
	\label{equa3}
	\begin{aligned}
		& {{h}_{t}}=\bar{A}{{h}_{t-1}}+\bar{B}{{x}_{t}} \\ 
		& {{y}_{t}}=\bar{C}{{h}_{t}} \\ 
	\end{aligned}
\end{equation}
where $t$ is the number of the input sequence, matrices $\bar{A}$, $\bar{B}$, and $\bar{C}$ represent the discrete parameters of the model, which include the sampling step $\Delta$. The $h_{t}$, $x_{t}$ and $y_{t}$ represent the hidden state, input and output of the system, respectively. 

The S6 \cite{gu2023mamba} is an adaptation to the S4 models \cite{gu2021efficiently}, where it makes the matrices $\bar{B}$, $\bar{C}$ and the sampling step $\Delta$ dependent on the input. This dependency is derived from the incorporation of the sequence length and batch size of the input, which enables the dynamic adjustment of these matrices for each input token. In this manner, S6 enables $\bar{B}$ and $\bar{C}$ to exert a dynamic influence on the state transition contingent on the input, thereby augmenting the model's content-awareness.

\subsection{Framework Overview}\label{Framework Overview}
Fig. \ref{fig_overall_framework}(a) depicts the pipeline of our FMOcc. As denoted in Fig. \ref{fig_overall_framework}(a), the input data for FMOcc consists of surround-view images, while the output os 3D semantic occupancy prediction results. FMOcc can be compartmentalized into four modules.

\textbf{2D image encoder.} The 2D image encoder will extract the surround-view features from surround-view images. For the 2D image encoder, we feed the surround-view images $\mathbf{I}$ to ResNet-based image encoder \cite{he2016deep}. The surround-view feature ${{\mathbf{F}}_{c}}\in {{\mathbb{R}}^{N\times H\times W\times C_{c}}}$ is the output of 2D image encoder. Here, $N$ is the number of cameras. $W$, $H$ are the spatial dimensions of the surround-view feature. $C_{c}$ is the channel dimension of the surround-view feature.

\textbf{2D-to-3D view transformation.} View transformation is a pivotal stage in the process of 3D perception, whereby the surround-view feature is converted into voxel representation. For view transformation, an auxiliary depth net is used to predict the depth distribution $\mathbf{D}\in {{\mathbb{R}}^{H\times W\times {{C}_{d}}}}$ of each pixel in the images, where $C_{d}$ denotes the number of depth bins. With the surround-view feature $\mathbf{F}_{c}$, the outer product $\mathbf{D}\otimes{{\mathbf{F}}_{c}}$ is employed to elevate pixels into grid-like pseudo-LiDAR points ${{\mathbb{R}}^{H\times W\times {{C}_{d}}\times {{C}_{c}}}}$, situated in camera coordinates. Subsequently, the pseudo points are transformed into world coordinates, warped into new voxel grids of fixed resolution $X\times Y\times Z$. Finally, 3D voxel-pooling is conducted to generate the 3D voxel feature $\mathbf{V}\in {{\mathbb{R}}^{X\times Y\times Z\times C_{c}}}$.

\textbf{Flow Matching SSM module.} Our proposed Flow Matching SSM module (FMSSM) is used to enhance the voxel feature $\mathbf{V_{input}}$ with 3D occupancy labels $Y$ to obtain the enhanced voxel feature ${\mathbf{V}_{output}}\in{{\mathbb{R}}^{X\times Y\times Z\times C_{c}}}$. The details of FMSSM will be introduced in Section \ref{Flow Matching SSM Module}.

\textbf{3D occupancy head.} After the process of FMSSM, we utilize a Multi-Layer Perceptron (MLP) to classify the category of each voxel grid. The process of 3D occupanct head is presented in the following equation:
\begin{equation}
	\label{equa4}
	{{\mathbf{O}}_{occ}}=MLP({{\mathbf{V}}_{e}})
\end{equation} 
\begin{figure*}
	\centering 
	\includegraphics[width=\textwidth]{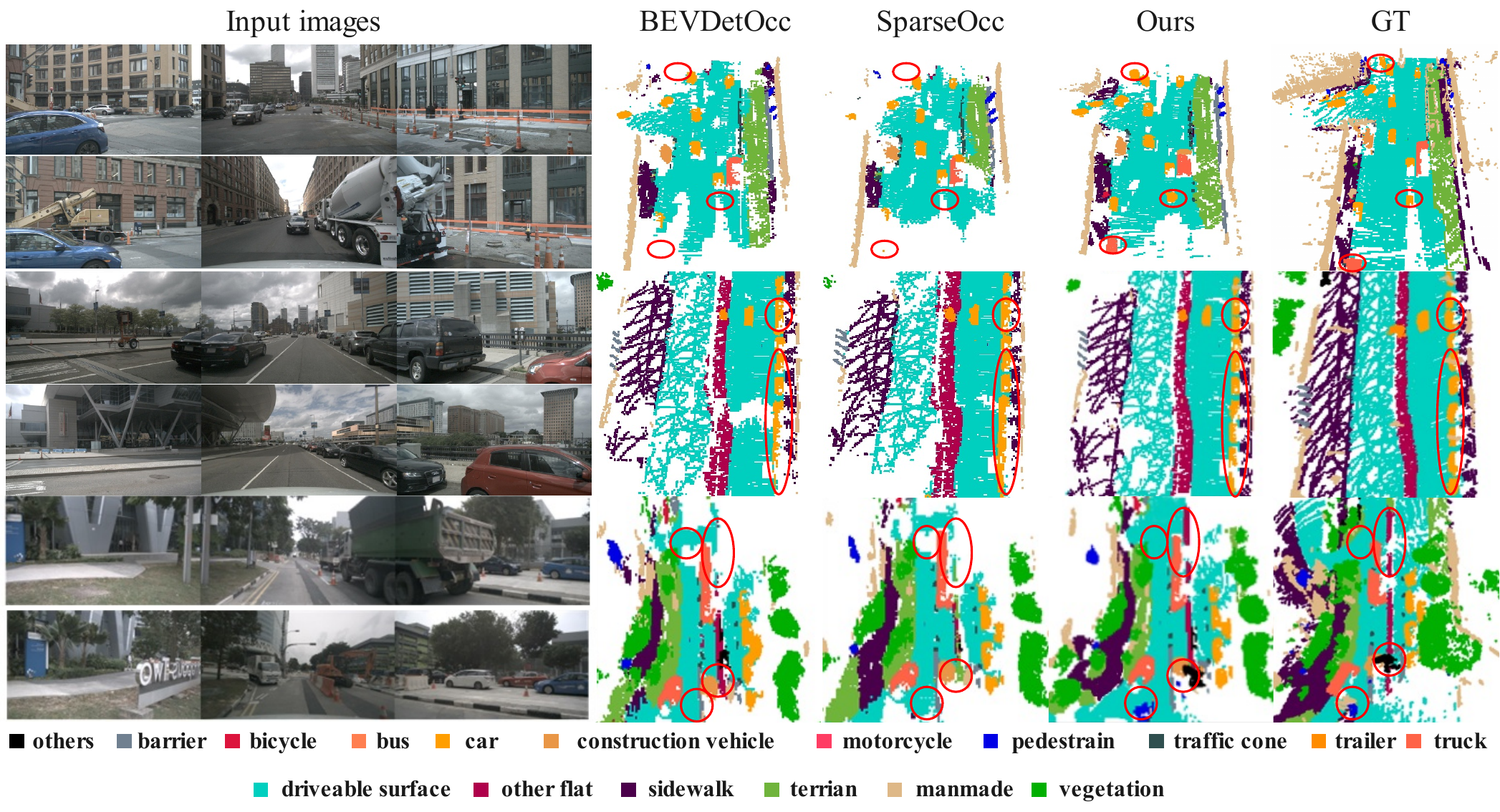}
	\caption{Visual comparison of our FMOcc with state-of-the-art methods on Occ3D-nuScenes. Compared to BEVDetOcc and SparseOcc, our FMOcc generates more precise prediction results of occlusion cases and long range cases (labeled in red circles).}
	\label{sota}
\end{figure*}

\begin{figure}
	\centering 
	\includegraphics[width=0.5\linewidth]{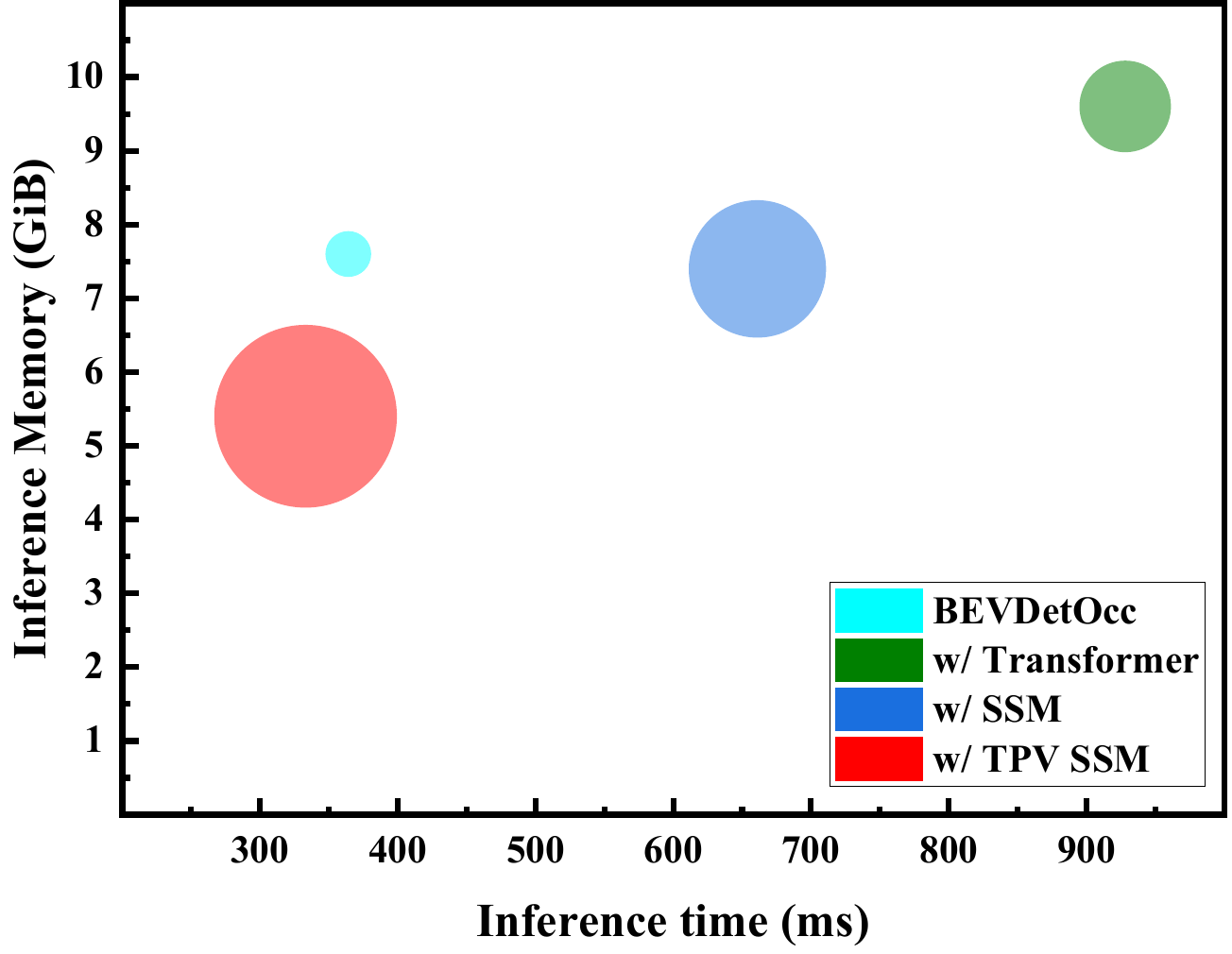}
	\caption{Demonstrating the efficiency of our proposed TPV SSM module in terms of inference memory and inference time, as well as its accuracy in mIoU. The size of the circles represents the value of mIoU.}
	\label{abla_tpv_ssm}
\end{figure}

\begin{figure}
	\centering 
	\includegraphics[width=0.5\linewidth]{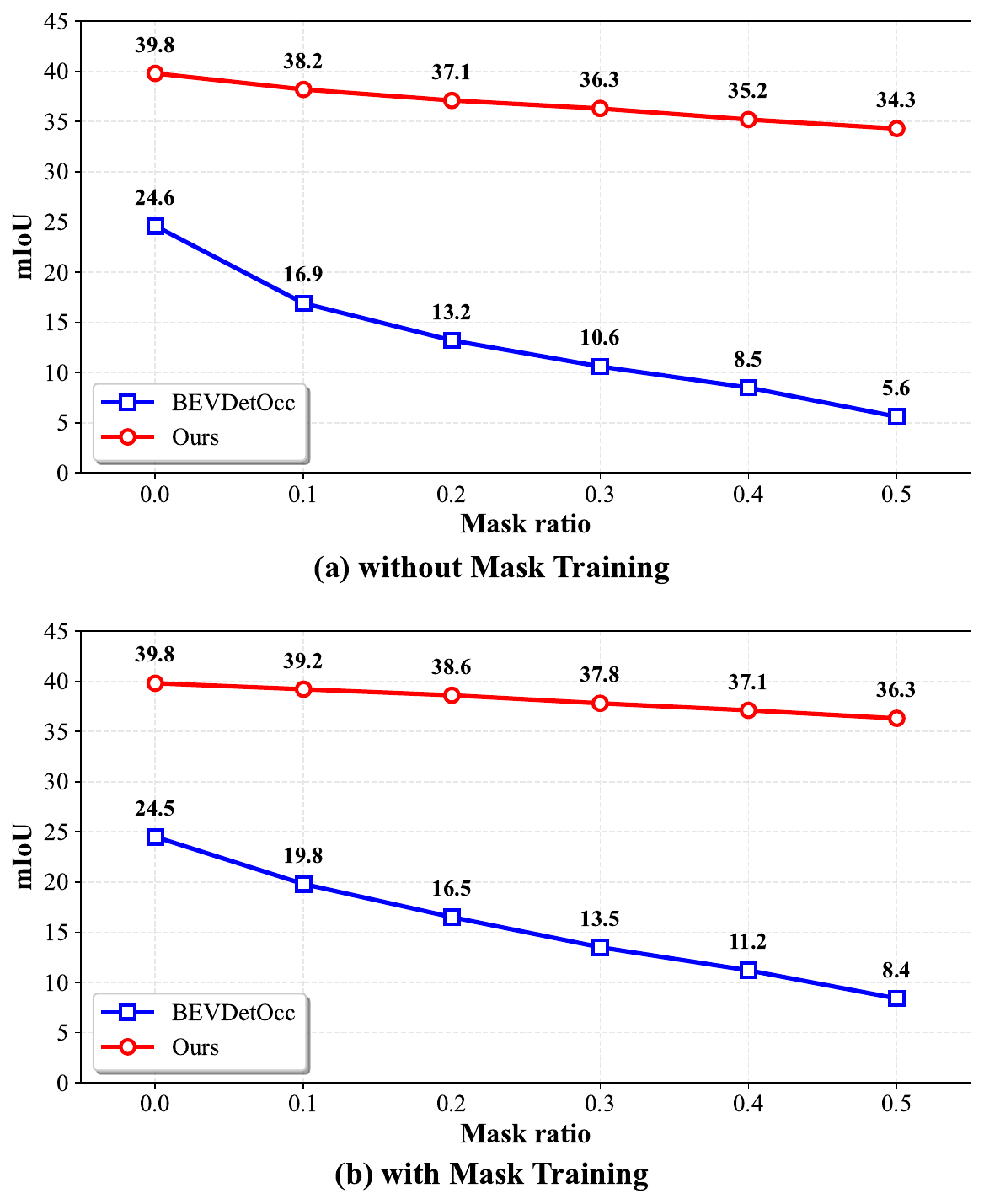}
	\caption{The robustness of our proposed FMOcc and the training strategy MT,  with the mask ratio ranging from 0 to 0.5. (a) Shows the mIoU of FMOcc and BEVDetOcc without MT. (b)Shows the mIoU of FMOcc and BEVDetOcc with MT.}
	\label{abla_mask_training}
\end{figure}

\begin{figure*}
	\centering 
	\includegraphics[width=\textwidth]{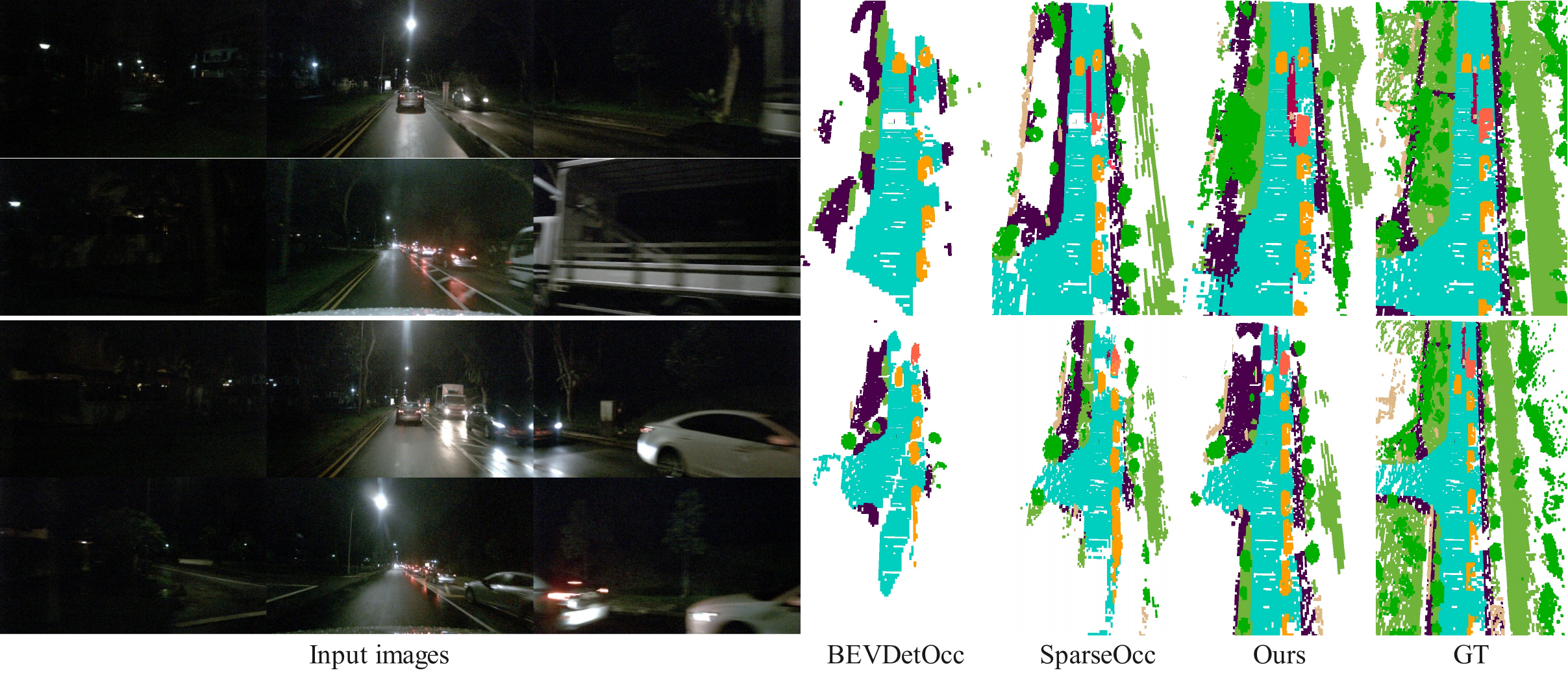}
	\caption{Visualizations for low-light environments on Occ3D-nuScenes validation set.}
	\label{abla_low_light}
\end{figure*}

\begin{figure*}
	\centering 
	\includegraphics[width=\textwidth]{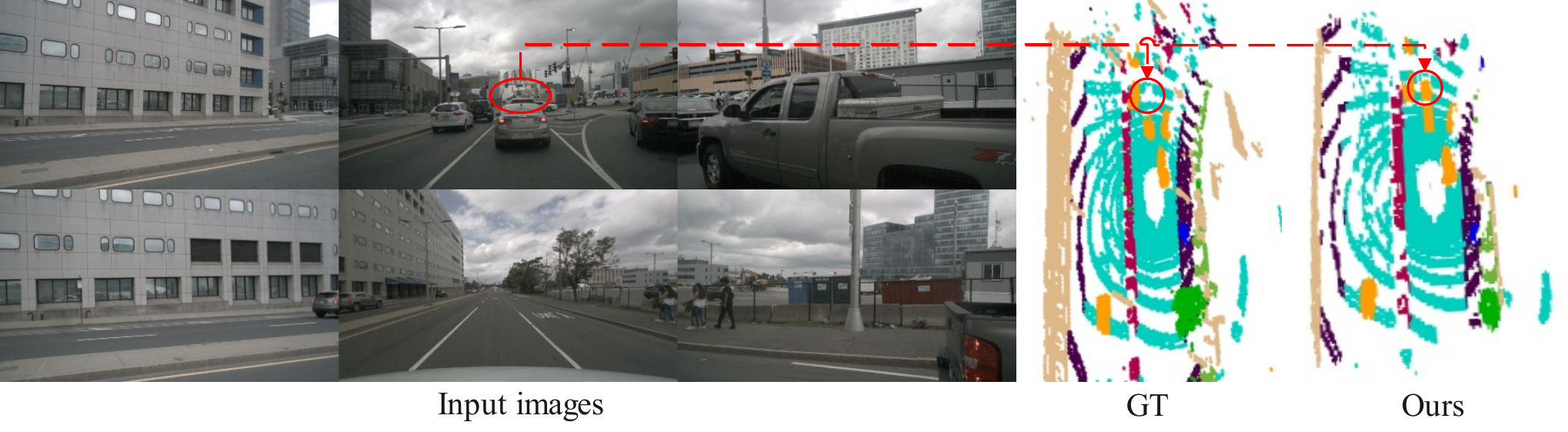}
	\caption{Visualizations for the miss groundtruth scene on Occ3D-nuScenes validation set. The red circle reprsents the missing "car" in the groundtruth.}
	\label{abla_lack_gt}
\end{figure*}
\subsection{Flow Matching SSM Module}\label{Flow Matching SSM Module}
As discussed above, most existing methods address the prediction of occluded objects by fusing voxel features from historical frames to generate combined voxel features. However, introducing temporal features will incur additional computational overhead. Furthermore, existing methods cannot effectively predict 3D occupancy in harsh environments, such as low-light conditions, or when sensor data is missing. Therefore, we propose Flow Matching SSM module (FMSSM) to generate the missing information in voxel $V_{input}$ due to few-frame input.

Figs. \ref{fig_overall_framework}(b) illustrates the entire process of FMSSM. The FMSSM begins with the refinement of semantic occupancy labels $Y$, involving encoding to produce informative target maps of flow matching, $V_{1}$. We first encode the labels $Y$:
\begin{equation}
	V_{1}=(sigmoid(embedding(Y))^{2}-1)\ast scale
\end{equation}
 The other input voxel $V_{input}$ is transformed into masked voxel features $V_0$ through MT. Subsequently, we determine the Optimal Transport Path (OTP) based on linear interpolation, thereby obtaining the input $V_t$ at time $t$: 
\begin{equation}
\label{eqa_opt}
	V_{t}=tV_{1} + (1-t)V_{0}
\end{equation}

After obtaining $V_{t}$, we carry out mean operations along the $x-axis$, $y-axis$ and $z-axis$ of the voxel features \cite{huang2023tri}. This process allows us to ultimately obtain tri-perspective view features at step $t$:
\begin{equation}
\label{eqa_tpv}
	T_{t}=\left [ T_{t}^{HW},T_{t}^{DH},T_{t}^{WD} \right ] = Mean(V_{t})
\end{equation}
where $T_{t}^{HW}$,$T_{t}^{DH}$,$T_{t}^{WD}$ represent the top, side and front views of a 3D scene at step $t$, respectively. Then, we take $T_t$ and step $t$ as the input of flow matching process, and get the velocity tpv feature $T_{t}^{v}$ after $t$ steps of flow matching:
\begin{equation}
	T_{t}^{v}=f_{\theta }(T_{t}, t)
\end{equation}
where $f_{\theta }$ is our proposed TPV-SSM layer. We will aggregate the tpv feature $T_{t}^{v}$ to get the velocity voxel feature $V_{t}^{v}$ at $t$:
\begin{equation}
	V_{t}^{v}=A(T_{t}^{HW}, T_{t}^{DH},T_{t}^{WD} )
\end{equation}
where $A$ is the summation operation. Furthermore, we incorporate Euler Integration Updates to compute the voxels at $t + \Delta t$ and ultimately derive $V_{output}$:
\begin{equation}
	V_{output}=V_{t} + \Delta t \times V_{t}^{v}
\end{equation}
Finally, we optimize the flow matching process by computing the loss between the predicted voxel velocity field $V_{t}^{V}$ and the target velocity field $x_{1}-x_{0}$:
\begin{equation}
	Loss_{flow}=MSE(V_{t}^{V}, (x_{1}-x_{0}))
\end{equation}

\subsection{TPV-SSM Layer}\label{TPV-SSM}
Due to the abundance of air features in 3D space, utilizing a Transformer-based method as $f_{\theta}$ in the diffusion model to predict $T_{t-1}$ leads to a waste of computational resources. Furthermore, the influence of the air feature on non-air feature can result in poor performance for long-distance prediction. Therefore, we propose the TPV-SSM layer based on S6 to process the input TPV features $T$. Figs. \ref{tpv_ssm_layers}(a) shows the overall process of TPV-SSM layer. We utilize three parallel architectures similar to Transformer encoders, but with the self-attention mechanism replaced by our proposed mechanism, denoted as Plane SSM blocks.

Figs. \ref{tpv_ssm_layers}(b) demonstrates the Plane SSM blocks of three planes. Figs. \ref{tpv_ssm_layers}(c) shows the core component of Plane SSM blocks, which is called Plane Selective SSM (PS$^{3}$M). First, we unfold the input plane features along four directions to obtain sequence features in those four directions. We define the sequence features as $x$. Then, the $x$ is processed to produce the $y$. The process can be defined as follows:
\begin{equation}
\begin{aligned}
	&h_{k}=e^{\Delta A}h_{k-1}+\Delta Bx_{k},\\
	&y_{k}=Ch_{k}+Dx_{k},\\
	&y=[y_{1},y_{2},y_{3},...,y_{L}],\\ 
\end{aligned} 
\end{equation}
where the $k$ represents the position of an element within the sequence. The $L$ is the total length of the sequence $x$. The $\Delta,B,C$ is produced by:
\begin{equation}
\begin{aligned}
	&B=s_{B}(x),\\
	&C=s_{C}(x),\\
	&\Delta =s_{\Delta}(x),\\
\end{aligned} 
\end{equation}
where $s_{B}(x), s_{C}(x), s_{\Delta}(x)$ are projection layers to get $B,C,\Delta$ from $x$. Based on the proposed $PS^{3}M$, we can eliminate redundant features in the input TPV features with linear time complexity, enhancing the ability of our proposed method to extract global contextual information. Therefore, our proposed method not only reduces the time required for multi-step inference of the diffusion model and the GPU memory footprint, but also significantly improves the prediction accuracy of distant 3D occupancy.

\subsection{Mask Training and Inference}\label{Mask Training}
In the task of 3D occupancy prediction for autonomous driving, the reliability of surround-view cameras is crucial. Acknowledging this, we introduce the Mask Training (MT) approach, aimed at bolstering the resilience and flexibility of our FMOcc, especially in situations where camera data might be obstructed or unavailable.

To emulate sensor failure or malfunction scenarios, we gradually elevate the dropout rate of camera inputs throughout training, ranging from 0 up to a preset cap of $\beta = 25$ . This can be expressed as:
\begin{equation}
\label{equa_mt}
V_{0}=V_{input}\odot  Bernoulli(\frac{\beta }{100}\cdot  \frac{e}{E} )
\end{equation}
where $e$ represents the current training epoch and $E$ is the total training epochs. $V_{input}$ is the voxel features extracted from surround-view images. $V_{0}$ refers to the voxel feature that is obtained by applying a mask, which is based on the Bernoulli distribution, to the original voxel feature $V_{input}$. $V_{0}$ is created by selectively hiding or revealing parts of $V_{input}$ according to a probabilistic mask generated by the Bernoulli distribution. This process allows for the generation of masked voxel feature, which can be useful training our proposed model with incomplete data.

\begin{table*}
	\belowrulesep=0pt
	\aboverulesep=0pt
	\centering     
         \resizebox{\linewidth}{!}{
	\begin{threeparttable}
		\caption{3D Occupancy Prediction Performence on Occ3D-nuScenes Dataset. Bold and Underline Denote the Best Performence and the Second-Best Performence, Respectively. Vis. Mask Means whether Models are Trained with Visible Masks.}\label{tbl1}
		\renewcommand{\arraystretch}{1.5}
           
		\begin{tabular}{c|c|c|c|c|c|c|c|c|c}
			\toprule
			Methods & Backbone & Input Size & Epochs & Vis. Mask & RayIoU & $\text{RayIoU}_{1m}$  & $\text{RayIoU}_{2m}$ & $\text{RayIoU}_{4m}$ & mIoU \\
			\hline
			MonoScene & R101  & 1600$\times$928  & 24 & $\surd$ & - & - & - & - & 6.0\\
			OccFormer & R101  & 1600$\times$928  & 24 & $\surd$ & - & - & - & - & 21.9\\
			TPVFormer & R101  & 1600$\times$928  & 24 & $\surd$ & - & - & - & - & 27.8\\
			CTF-Occ & R101  & 1600$\times$928  & 24 & $\surd$ & - & - & - & - & 28.5\\
			RenderOcc & SwinB & 1408$\times$512  & 24 & $\surd$ & - & - & - & - & 26.1\\
			BEVFormer & R101 & 1600$\times$900 & 24 & $\surd$ & 32.4 & 26.1 & 32.9 & 38.0 & 39.3\\
			FB-Occ (16f) & R50 & 704$\times$256 & 24 & $\surd$ & 33.5 & 26.7 & 34.1 & 39.7 & 39.1\\
			FlashOcc & R50 & 704$\times$256 & 24 & $\surd$ & - & - & - & - & 32.0\\
			COTR &R50&704$\times$256&24& $\surd$&41.0&\underline{36.3}&41.7&45.1&\underline{44.5}\\
			OPUS-L &R50&704$\times$256 &100&$\surd$&\underline{41.2}&34.7&\underline{42.1}&\underline{46.7}&36.2\\
			FMOcc (2f)&R50&704$\times$256&24&$\surd$&\textbf{46.5}&\textbf{40.3}&\textbf{47.7}&\textbf{51.5}&\textbf{47.9}\\
			\hline
			BEVFormer & R101 & 1600$\times$900 & 24 & - & 33.7 & - & - & - & 23.7\\
			FB-Occ (16f) & R50 & 704$\times$256 & 24 & - & 35.6 & - & - & - & 27.9\\
			RenderOcc&SwinB&1408$\times$512&12&-&19.5&13.4&19.6&25.5&-\\
			BEVDetOcc (2f)&R50&704$\times$256&90&-&29.6&23.6&30.0&35.1&24.6\\
			BEVDetOcc (8f)&R50&704$\times$384&90&-&32.6&26.6&33.1&38.2&26.3\\
			SparseOcc (8f)&R50&704$\times$256&24&-&34.0&28.0&34.7&39.4&30.6\\
			SparseOcc (16f)&R50&704$\times$256&24&-&35.1&29.1&35.8&40.3&30.9\\
			STCOcc &R50&704$\times$256 &36&$-$&\underline{41.7}&\underline{36.2}&\underline{42.7}&\underline{46.4}&\underline{35.6}\\
			\hline
			
			FMOcc (2f)&R50&704$\times$256&24&-&\textbf{43.1}&\textbf{36.9}&\textbf{44.3}&\textbf{48.1}&\textbf{39.8}\\
			\bottomrule
		\end{tabular}
	\end{threeparttable}}
\end{table*}

\begin{table}
	\belowrulesep=0pt
	\aboverulesep=0pt
	\centering
	      
	\begin{threeparttable}
		\caption{3D Occupancy Prediction Performence on OpenOcc Dataset. Bold and Underline Denote the Best Performence and the Second-Best Performence.C and L Denote Camera and Lidar Supervision. Mem Denote Training Memory.}\label{tbl2}
		\renewcommand{\arraystretch}{1.5}
		\begin{tabular}{c|ccc|cc}
			\toprule
			Methods  & Sup. & Backbone& Input Size & $\text{RayIoU}$ &Mem(G) \\
			\hline
			RenderOcc & L & R101 & 1600$\times$900 & 36.7 &-  \\
			OccNet & 3D & R101 & 1600$\times$900 & 39.7  &-\\
			BEVFormer & 3D & R50 & 1600$\times$900 & 28.1 & 26.0\\
			FB-Occ & 3D & R50 & 704$\times$256 & 32.3  &11.1\\
			SparseOcc & 3D & R50 & 704$\times$256 & 33.4 & 15.8\\
			STCOcc & 3D& R50 & 704$\times$256 & \underline{40.8}  & \textbf{8.7}\\
			\hline
			FMOcc (2f) &3D & R50 & 704$\times$256 & \textbf{42.6} & \underline{8.8}\\
			\bottomrule 
		\end{tabular}
	\end{threeparttable}
\end{table}

\begin{table}
	\belowrulesep=0pt
	\aboverulesep=0pt
	\centering
	      
	\begin{threeparttable}
		\caption{Ablation on the Proposed FMSSM Module.}\label{tbl_abla_fmssm}
		\renewcommand{\arraystretch}{1.5}
		\begin{tabular}{c|cccc}
			\toprule
			Methods  & RayIoU & $\text{RayIoU}_{1m}$ & $\text{RayIoU}_{2m}$ & $\text{RayIoU}_{4m}$ \\
			\hline
			w/o FMSSM & 32.6 & 26.6 & 33.1 & 38.2  \\
			w/ FMSSM  & 43.1 & 36.9 & 44.3 & 48.1  \\
			\bottomrule 
		\end{tabular}
	\end{threeparttable}
\end{table}

\begin{table}
	\belowrulesep=0pt
	\aboverulesep=0pt
	\centering
	\caption{Ablation on the Flow Matching vs. Diffusion Model.}\label{tbl_abla_fm_vs_dm}    
	\begin{threeparttable}
		\renewcommand{\arraystretch}{1.5}
        \setlength{\tabcolsep}{3pt} 
		
		\begin{tabular}{c|cccc|c}
			\toprule
			Methods  & RayIoU & $\text{RayIoU}_{1m}$ & $\text{RayIoU}_{2m}$ & $\text{RayIoU}_{4m}$ & Latency (ms)\\
			\hline
			w/o Generative Method & 32.6 & 26.6 & 33.1 & 38.2 & 388 \\
			Diffusion Model & 38.6 & 33.2 & 39.4 & 43.4 & 691 \\
			Flow Matching  & 43.1 & 36.9 & 44.3 & 48.1 & 333 \\
			\bottomrule 
		\end{tabular}
	\end{threeparttable}
\end{table}

\begin{table}
	\belowrulesep=0pt
	\aboverulesep=0pt
	\centering     
	
	\begin{threeparttable}
		\caption{Ablation on the Proposed TPV SSM Layer. Training Denotes Training Memory. Inference Denotes Inference Memory}\label{tbl_abla_tpvssm}
		\renewcommand{\arraystretch}{1.5}
		\begin{tabular}{c|cc|c|c}
			\toprule
			\multirow{2}{*}{Methods}  & \multirow{2}{*}{RayIoU}   & \multirow{2}{*}{mIoU} & \multicolumn{2}{c}{Memory Usage(GiB)}\\ \cline{4-5}
			&&& Training & Inference\\
			\hline
			w/ Transformer & 38.9 & 35.2 & 12.4 & 9.6\\
			w/ SSM  & 42.2  & 39.1 & 10.8 & 7.4\\
			w/ TPV SSM & 43.1  & 39.8 & 8.8 & 5.4\\
			\bottomrule
		\end{tabular}
	\end{threeparttable}
\end{table}

\begin{table}
	\belowrulesep=0pt
	\aboverulesep=0pt
	\centering 
	    
	\begin{threeparttable}
		\caption{Performance Comparison of The Different Ranges of Scenes between FMOcc and BEVDetOcc.}\label{tbl_abla_distant}
		\renewcommand{\arraystretch}{1.5}
		
		\begin{tabular}{c|ccc|ccc}
			\toprule
			\multirow{2}{*}{Methods}  & \multicolumn{3}{c|}{RayIoU} & \multicolumn{3}{c}{mIoU} \\\cline{2-4}\cline{5-7}
			& 25m & 50m &100m & 25m & 50m &100m \\ 
			\hline
			BEVDetOcc & 42.5 & 38.9 & 32.6 & 38.7 & 33.6 & 26.3 \\
			Ours  & 48.6 & 44.9 & 39.6 & 45.5 & 41.9 & 34.4 \\
			Improvements (\%) & $+$6.1 & $+$6.0 & $+$7.0 & $+$6.8 & $+$8.3 & $+$8.1\\
			\bottomrule
		\end{tabular}
	\end{threeparttable}
\end{table}
\begin{table}
	\belowrulesep=0pt
	\aboverulesep=0pt
	\centering     
	
	\begin{threeparttable}
		\caption{Ablation on the Different Backbone. Training Denotes Training Memory. Inference Denotes Inference Memory}\label{tbl_abla_backbone}
		\renewcommand{\arraystretch}{1.5}
		\begin{tabular}{c|cc|c|c}
			\toprule
			\multirow{2}{*}{Methods}  & \multirow{2}{*}{RayIoU}   & \multirow{2}{*}{mIoU} & \multicolumn{2}{c}{Memory Usage(GiB)}\\ \cline{4-5}
			&&& Training & Inference\\
			\hline
			w/ ResNet50  & 43.1  & 39.8 & 8.8 & 5.4\\
			w/ ResNet101  & 45.2  & 41.6 & 10.8 & 7.4\\
			w/ SwinB & 48.9 & 44.3 & 12.5 & 8.6\\
			\bottomrule
		\end{tabular}
	\end{threeparttable}
\end{table}
\section{Experiments}
In this section, we present the experimental setup and results obtained in this study, structured into three primary subsections for clarity. Section \ref{setting} details the dataset, evaluation metrics and training details employed in our experiments. In Section \ref{comparison sota}, we conduct a comparative analysis of our proposed FMOcc against several state-of-the-art 3D occupancy prediction approaches. Section \ref{ablation study} presents the ablation study, an essential part of our analysis that investigates the contribution of each component within our proposed framework.
\subsection{Setting}\label{setting}
\textbf{Dataset.} We conducted occupancy analysis on the Occ3D-nuScenes \cite{tian2024occ3d} and OpenOcc \cite{tong2023scene} datasets, utilizing a data collection vehicle equipped with a comprehensive sensor suite consisting of one LiDAR sensor, five radars, and six cameras. This setup provides a 360-degree view of the vehicle's surroundings. The dataset is divided into 700 scenes for training and 150 scenes for validation, with each scene spanning 20 seconds in duration. Ground truth labels are annotated every 0.5 seconds to ensure precise temporal resolution. The Occ3D-nuScenes and OpenOcc datasets cover an extensive spatial range, stretching from -40 meters to 40 meters along both the x and y axes, and from -1 meter to 5.4 meters along the z-axis. Occupancy labels are defined using voxels with dimensions of $0.4m \times 0.4m \times 0.4m$, encompassing 17 distinct categories on Occ3D-nuScenes and 16 distinct categories on OpenOcc. Each driving scene is annotated with perceptual data captured at a frequency of 2 Hz, providing a rich and detailed dataset for our occupancy analysis.

\textbf{Metrics.} The semantic segmentation performance is evaluated using both the mean intersection-over-union (mIoU) and RayIoU. The latter is employed to address the depth inconsistency penalty inherent in the conventional mIoU criteria \cite{liu2024fully}.

\textbf{Training details.}
In our paper, we utilize ResNet50 as the backbone architecture and feed images with resolutions of $704\times256$ into the model. Our experimental setup is implemented in PyTorch, with a total batch size of 8 distributed over 4 A40 GPUs. The training epoch is set to 24. We employ the AdamW \cite{loshchilov2017decoupled} optimizer, setting the weight decay to $1\times 10^{-2}$. The learning rate remains constant at $1\times 10^{-4}$, except for a linear warmup during the initial 200 iterations. All loss weights in our methodology are uniformly set to 1.0. Additionally, we have deliberately omitted the use of camera masks throughout the training phase of our model.

\subsection{Comparison with state-of-the-art (SOTA) methods}\label{comparison sota}
\subsubsection{Quantitative analysis}
To guarantee fair comparisons, all results have been either directly implemented by their respective authors or accurately reproduced by utilizing official codes.

We categorize these methods into two groups: with visible masks and without visible masks. The Occ3D-nuScenes dataset provides visible masks that represent voxel observations in the current views, determined via ray-casting. By integrating these masks into the training process, we significantly enhance the model's learning progress. However, this approach involves a trade-off, as it may cause the model to disregard areas beyond the visible region, such as obstacles and the sky, ultimately leading to a decrease in visualization quality \cite{pan2023uniocc}. Therefore, in our proposed FMOcc, we adopt a training approach without the use of visible masks. 

Table \ref{tbl1} shows the comparative results of our FMOcc against other state-of-the-art camera-based 3D occupancy prediction methods on the Occ3D-nuScenes dataset. We use BEVDetOcc as our baseline model. Table \ref{tbl1} shows that our FMOcc achieves better results with two-frame images than BEVDetOcc with 8-frame images. The improvement is $32.2\%$ and $51.3\%$ on metrics $RayIoU$ and $mIoU$, respectively. Compared to the state-of-the-art SparseOcc, our proposed FMOcc performs better when using two-frame images than SparseOcc with 8-frame images, and is also better than SparseOcc with 16-frame images. The improvements of $RayIoU$ and $mIoU$ compared to SparseOcc(16f) are $22.7\%$ and $28.8\%$. When adopting the visible mask during training, our proposed model demonstrates improvements of $11.7\%$ and $3.3\%$ in mIoU and RayIoU , respectively, compared to the state-of-the-art STCOcc model.

Table \ref{tbl2} demonstrates the superiority of our proposed method on the OpenOcc dataset. Compared with STCOcc, our method achieves a 4.4\% improvement in the RayIoU metric without increasing memory consumption. Experimental results further validate that the proposed approach exhibits favorable performance across diverse datasets.

\subsubsection{Qualitative analysis}
Our method is compared with two high-performance benchmarks, BEVDet4D and SparseOcc, as illustrated in Figs. \ref{sota}. In Fig. \ref{sota}, we highlight the advantages of our proposed method compared to the other two methods with red circles. In the third row of Fig. \ref{sota}, the visualization vividly demonstrates the efficacy of our proposed method in predicting the \textit{"sidewalk"} regions that are otherwise obscured by \textit{"construction vehicle"}. To elaborate on the visualization in Fig. \ref{sota}, not only does our proposed method effectively predict the \textit{"sidewalk"} regions obscured by \textit{construction vehicle}, but it also enhances the accuracy of identifying distant drivable areas. This improvement is evident in the clearer and more precise delineation of road boundaries and drivable spaces, even at greater distances from the ego car's perspective. The first and second rows in Fig. \ref{sota} demonstrate the improvement of our proposed method in predicting \textit{"car"} under occlusion conditions and at long distances.

\subsection{Ablation study}\label{ablation study}
In this section, we conduct an ablation study to dissect the contributions of critical components within the FMOcc architecture. We specifically examine the influence of the essential modules: FMSSM, TPV-SSM and mask training. Through this systematic ablation studies, we evaluate the individual contributions of each module integrated into the FMOcc framework via these ablation studies.

\subsubsection{FMSSM module} 
Table \ref{tbl_abla_fmssm} demonstrates the effectiveness of our proposed FMSSM module, which can improve our proposed method by 32.2\% in RayIoU. As the threshold for RayIoU becomes smaller, the improvement in prediction accuracy achieved by our proposed FMSSM module becomes more significant. This proves that under stricter metrics, the proposed FMSSM module can still effectively enhance prediction results. 

Table \ref{tbl_abla_fm_vs_dm} compares the performance of methods based on flow matching, diffusion models, and non-generative approaches. Although non-generative methods exhibit faster inference speeds, their prediction accuracy is inferior to generative methods. While diffusion models enhance prediction accuracy, they significantly compromise inference speed. Our proposed FMOcc, leveraging generative methods based on flow matching, not only improves prediction accuracy but also maintains remarkable competitiveness in inference speed.

\subsubsection{TPV-SSM layer}
Table. \ref{tbl_abla_tpvssm} demonstrates the substantial enhancement in 3D occupancy prediction accuracy achieved by our proposed TPV-SSM layer, while simultaneously highlighting its superior efficiency in memory utilization compared to Transformer and SSM layer. Through comparison, it is evident that the TPV-SSM layer not only significantly improves 3D occupancy prediction accuracy but also excels in memory management, achieving a dual optimization of computational resources and prediction performance. 

Fig. \ref{abla_tpv_ssm} clearly illustrates the advantages of our proposed TPV-SSM layer in terms of accuracy, memory usage, and inference speed. In the Fig. \ref{abla_tpv_ssm}, the size of the circles represents prediction accuracy, with larger circles indicating higher accuracy. Circles positioned closer to the bottom-left corner signify more efficient models in terms of both inference speed and memory usage. As shown in Fig. \ref{abla_tpv_ssm}, TPV-SSM layer significantly outperforms BEVDetOcc in accuracy and is slightly faster in inference speed.

\subsubsection{Mask training} 
Fig. \ref{abla_mask_training} analyzes the contribution of mask training proposed in this paper on the robustness of the model. Fig. \ref{abla_mask_training}(a) and Fig. \ref{abla_mask_training}(b) represent two scenarios: the case without employing the mask training strategy and the case where the mask training strategy is applied, respectively. Mask ratio represents the proportion of features that are corrupted during the inference process. According to Fig. \ref{abla_mask_training}, the mask training proposed in this paper can effectively improve the prediction performance of BEVDetOcc and the proposed FMOcc model when features are corrupted. Even without employing of the mask training strategy, our proposed FMOcc demonstrates a remarkable ability to maintain high prediction accuracy when faced with a 50\% corruption of features.

\subsubsection{Performance of distant prediction} 
Table \ref{tbl_abla_distant}presents a comparison of the prediction performance between our proposed method and BEVDetOcc at three distances: 25m, 50m, and 100m. The results demonstrate that our model achieves significant improvements across all distances. Specifically, the RayIoU metric is enhanced by 6.1\%, 6.0\%, and 7.0\% at 25m, 50m, and 100m respectively, which fully attests to the superior performance of our proposed method in short, medium, and long-distance predictions. Similarly, the mIoU metric also experiences increases of 6.8\%, 8.3\%, and 8.1\% at these respective distances, further validating the comprehensive improvement in prediction accuracy achieved by our model. In summary, our proposed method exhibits outstanding performance in predictions at short, medium, and long distances, demonstrating a significant advantage over BEVDetOcc.

\subsubsection{Different Backbone}
Table \ref{tbl_abla_backbone} compares the prediction performance of our model with different backbones. The data in the Table \ref{tbl_abla_backbone} demonstrate that the proposed method exhibits superior prediction performance across various backbones. Notably, even when using the Swin Transformer base model, it effectively controls the memory usage during both training and inference, further highlighting the efficiency of the proposed model.

\subsection{Visualization on Low-light Scenes}
To further demonstrate the resilience of our approach in low-light conditions, we provide 
additional visual results. In the first row of Fig. \ref{abla_low_light}, our proposed method can effectively predict distant \textit{"car"} and nearby \textit{"truck"} under low-light conditions, and it can also generate the \textit{"driveable surface"} effectively. The second row of Fig. \ref{abla_low_light} demonstrates that our proposed method can not only predict moving objects effectively under low-light conditions but also accurately predict certain road structures, such as \textit{"sidewalk"} and \textit{"driveable surface"}.

\subsection{Visualization of Scenes with Missing Labels}
Fig. \ref{abla_lack_gt} demonstrates the effectiveness of our proposed method in scenarios with missing labels. As can be seen from the input image in Fig. \ref{abla_lack_gt} the object marked with a red box exists in the actual input image, but the label has omitted this object. In this challenging situation, our proposed method still accurately predicts the object that is occluded in front.

\section{Conclusion}
This work focus on camera-based 3D semantic occupancy prediction. Existing methods always rely on fusing historical frame information to predict occluded scenes and distant scenes. However, fusing historical frames requires additional historical data and results in significant computational resource consumption. Therefore, we propose the two-frame FMOcc to generate the occluded features and distant features based on the refine TPV features. We first designed a feature refinement module based on diffusion model to generate the missing features. Then, we design the TPV SSM layer as the $f_{\theta}$ of diffusion model which can improve the efficiency of the model and the ability of predicting distant scenes. Finally, the Mask Training (MT) is proposed to enhance the robustness of the FMOcc. Extensive experiments on the Occ3D-nuScenes benchmark demonstrate the effectiveness of our FMOcc, which achieves state-of-the-art performance while using only two frame.

In the future, we hope to generate future 3D occupancy scenes for dynamic objects such as "car" and static scenes such as "driveable area", based on world model combined with occupancy flow prediction and occupancy prediction.

\bibliographystyle{unsrt}  


\end{document}